\def\year{2019}\relax
\documentclass[letterpaper]{article} 
\usepackage{aaai19}  
\usepackage{times}  
\usepackage{helvet}  
\usepackage{courier}  
\usepackage{url}  
\usepackage{graphicx}  
\usepackage{amsmath}
\usepackage{algorithm}
\usepackage{multirow}
\usepackage{subfig}
\usepackage[noend]{algpseudocode}
\usepackage{booktabs}
\usepackage{xcolor}
\usepackage{makecell}
\begin{document}
%
\title{Collaborative Deep Learning Across Multiple Data Centers}
\author{Kele Xu$^{1,2}$, Haibo Mi$^{1,2}$, Dawei Feng$^{1,2}$, Huaimin Wang$^{1,2}$, Chuan Chen$^3$, Zibin Zheng$^3$, Xu Lan$^4$\\
$^1$ National Key Laboratory of Parallel and Distributed Processing, Changsha, China\\
$^2$ College of Computer, National University of Defense Technology, Changsha, China\\
$^3$ School of Data and Computer Science, Sun Yat-Sen University, Guangzhou, China\\
$^4$ Queen Mary University of London, London, UK\\}

\maketitle
\begin{abstract}

Valuable training data is often owned by independent organizations and located in multiple data centers. Most deep learning approaches require to centralize the multi-datacenter data for performance purpose. In practice, however, it is often infeasible to transfer all data to a centralized data center due to not only bandwidth limitation but also the constraints of privacy regulations. Model averaging is a conventional choice for data parallelized training, 
but its ineffectiveness is claimed by previous studies as deep neural networks are often non-convex. In this paper, we argue that model averaging can be effective in the decentralized environment by using two strategies, namely, the cyclical learning rate and the increased number of epochs for local model training. With the two strategies, we show that model averaging can provide competitive performance in the decentralized mode compared to the data-centralized one. In a practical environment with multiple data centers, we conduct extensive experiments using state-of-the-art deep network architectures on different types of data. Results demonstrate the effectiveness and robustness of the proposed method.

\end{abstract}

\section{Introduction}

The sensitive data, such as medical imaging data, genetic sequences, financial records and other personal information, is often managed by independent organizations like hospitals and companies \cite{tian2016}. 
%
Many deep learning (DL) algorithms prefer to 
use as much data as possible distributed in different organizations for training,
because the performance of these DL algorithms directly depends on the amount of high-quality data
not only for rarely occurring patterns but also for the robustness to the outliers \cite{amir2017}.
%
In practice, however,
directly sharing data between different organizations is of great difficulties due to many reasons 
including privacy protection, legal risk consideration and conflict of interests.
Therefore,
it has become an important research topic for both academy and industry
to fully employ the data of different organizations 
for training DL models without centralizing the data,
while achieving similar performance compared to centralized training after moving all data together.


Recently, there has been a trend to use collaborative solvers to train a global model
on geo-distributed, multi-datacenter data
without directly sharing data between different data centers \cite{cano2016towards,hsieh2017}.
%
Specifically, 
several participants independently train the DL models for a while, 
and periodically aggregate their local updates to construct a shared model. 
Only parameters are exchanged and all the training data is kept in the original places \cite{mcmahan2016}. 
However, there are several challenges for this approach:

\begin{itemize}
	\item Large performance gap compared to the centralized mode:
	When training on the disjoint multi-party data, traditional deep models using Stochastic Gradient Descent (SGD) are difficult to provide competitive performance compared to their centralized mode. Further, with limited data size, the local learner is vulnerable to fall into the local optima, as deep models are generally non-convex.
	
	\item High communication cost: 
	different datasets are stored on different data centers (on private cloud or public cloud). DL algorithms typically require frequent communication to exchange parameter updates such that the shared deep model is of superior performance. However, current parameter servers are designed for high-speed local area networks (LANs). Due to the limitation of network bandwidth of wide-area networks (WANs), parameters of the global model cannot be exchanged frequently in the multi-datacenter environment. Therefore, it is necessary to decrease the communication cost for parameter exchange between different data centers, while retaining the accuracy of the shared model.
	
	\item High model aggregation complexity: The update strategy to aggregate the local models is complicated. As the different participant has its own training setting, the approach to aggregate local learners should be simple. In addition, the aggregation method should support the learning procedure using different deep neural network architectures.
\end{itemize}

In this work, we propose a multi-datacenter based collaborative deep learning method (denoted as co-learning), which (1) minimizes the performance gap between the centralized and decentralized modes, (2) minimizes the inter-datacenter communication cost during the co-training procedure over WANs, 
(3) is applicable to a wide variety of deep network architectures without any change.

The co-learning approach proposes two strategies to improve the performance of a shared model in distributed learning, based on the conventional model averaging method.
First, we adopt the modified cyclical learning rate \cite{izmailov2018averaging}, 
so as to avoid falling into the local optima during the local training procedure.
Second, we enlarge the number of local epochs when the difference between two consecutive shared models decreases to be less than a threshold,
so as to increase the diversity between local models and reduce the inter-datacenter communication cost. 
The synchronization period is extended from milliseconds or seconds to ten of minutes or even hours. 

Surprisingly, 
despite the claims from previous studies \cite{povey2014parallel,mcmahan2016}, 
we find that model averaging in the decentralized mode can provide competitive performance compared to the traditional centralized mode.
Extensive experiments are conducted on three different tasks: image classification, text classification and audio classification. Using the co-learning method, we have tested various state-of-the-art neural network architectures including VGGNet \cite{simonyan2014very}, ResNet \cite{he2016deep}, DenseNet \cite{huang2017densely} and Capsule architectures \cite{sabour2017dynamic}. All the experiments reveal that the proposed co-learning approach can provide superior performance in the decentralized mode. In summary, the main contributions include:

\begin{itemize}
	\item We propose a collaborative deep learning approach using model averaging. 
	With two simple strategies 
	(cyclical learning rate and increased number of local training epochs), 
	we show that model averaging can provide competitive performance compared to the centralized mode.
	
	\item Our approach enables the training of collaborative deep learning in the practical WAN environment. 
	
	\item  The proposed co-learning is flexible enough to be applied to a wide range of deep learning architectures without any change.
\end{itemize}

The remainder of this paper is organized as follows. Section 2 descries the related work, while Section 3 presents the details of our co-learning approach. Section 4 describes the experimental results, the discussion and conclusion are given in Section 5.

\section{Related Work}

With the increase of data size and model complexity, training a deep neural network can take long time. An increasing trend to scale deep learning is to partition the training dataset, concurrently train separate models on the disjoint subset. By aggregating the updates of local model's parameters via a parameter server, a shared model can be constructed. In this paper, we define this method as collaborative deep learning, which can be applied in the practical situation where each participant wants to hide their own training data from each other.

\subsection{Parallelized Stochastic Gradient Descent}
Many recent attempts have been made to parallelized SGD based learning schemes across multiple data centers \cite{hsieh2017,zhang2017poseidon}. Nevertheless, the geo-distributed nature of data prevents its widespread utilization between organizations, due to the aforementioned reasons like limitations in cross data center connectivity, or data sovereignty regulations restriction. To break through these restrictions, increasing effort has been made. \cite{shokri2015privacy} uses the parallel stochastic gradient descent algorithm to train the model for the consideration of privacy preservation. The communication cost between the client and the server is prohibitively high, thereby can seldom be deployed in WAN scenarios due to the bandwidth limit. \cite{tian2016} proposed a secure multi-party computation (MPC) approach for simple and effective computations, yet its overhead for complex computations and the model training is nontrivial. Consequently, this approach is more suitable for shallow ML models, while it is difficult to be applied to deep learning models \cite{zinkevich2010parallelized}.

Furthermore, to reduce the communication cost, many compression approaches have been explored, such as, gradient quantization \cite{alistarh2017qsgd} and network pruning \cite{lin2017deep}, knowledge distillation \cite{anil2018large,hinton2015distilling}.

\subsection{Model averaging}
For collaborative deep learning, model averaging is an alternative method for parallelized SGD  \cite{su2015experiments,povey2014parallel}. However, most of the previous literatures \cite{sun2017ensemble,goodfellow2014qualitatively} claimed that traditional model averaging cannot provide satisfied performance in the distributed setting, as a deep neural network is a highly non-convex model. For example, \cite{povey2014parallel} claimed that the model averaging algorithm did not work well for speech recognition models. The main reason to support these claims was that: when the size of the data is limited for the training of a local model, the local models may fall into different local-optima. The shared model obtained by averaging the local model's parameters, might even perform worse than any local model. Moreover, in the follow-up step, the shared model would be used as a new starting point of the successive iterations of local training, and the poor performance of the shared model would drastically slow down the convergence of the training process and further decreased the performance of the shared model \cite{sun2017ensemble}. To avoid falling into local optima, many regularization methods were proposed \cite{srivastava2014dropout,ioffe2015batch}. In \cite{izmailov2018averaging}, it was found that using a cyclical learning rate could lead to better generalization than the conventional training. 

A federated learning approach \cite{mcmahan2016} was proposed for a data parallelization in the context of deep learning. It targeted to solve the model training on massive mobile devices, and a fixed number of epochs for local model training was employed for the devices. However, We utilize a modified cyclical learning rate and an increasing number of epochs for local model training to get competitive performance in the decentralized mode with comparison to the centralized one.

\section{Methodology}

\subsection{Notation and problem formulation}
A typical process of parallel training for deep models is illustrated in Figure 1. Participants train their local models with the individual deep learning platform in their private data centers (in private clouds or trusted public clouds). These local data centers communicate over WANs. In the piratical situation, due to the limitation of WAN bandwidth, participants cannot exchange updates frequently.

In the following, we denote a deep neural network as $f(\bf{w})$, where ${\bf w}$ represents the parameters of this neural network model. In addition, we denote the outputs of the model $f(\bf{w})$ on the input $x$ as $f(\bf{w},x)$. In the parallel training of deep models, suppose there are $K$ participants and each of them holds a local dataset $D_k=\{(x_{k,1},y_{k,1}),...,(x_{k,m_k},y_{k,m_k})\}$ with size $m_k,~k\in\{1,...,K\}$. Denote the weight of the neural network model at the iteration $t$ of $i$-th round (with $T_i$ epochs been performed) on the participant $k$ as ${\bf w}_k^{i,t}$. Then a typical parallel training procedure for neural network implements the following two steps:

\begin{figure}
	\centering
	\includegraphics[scale=0.6]{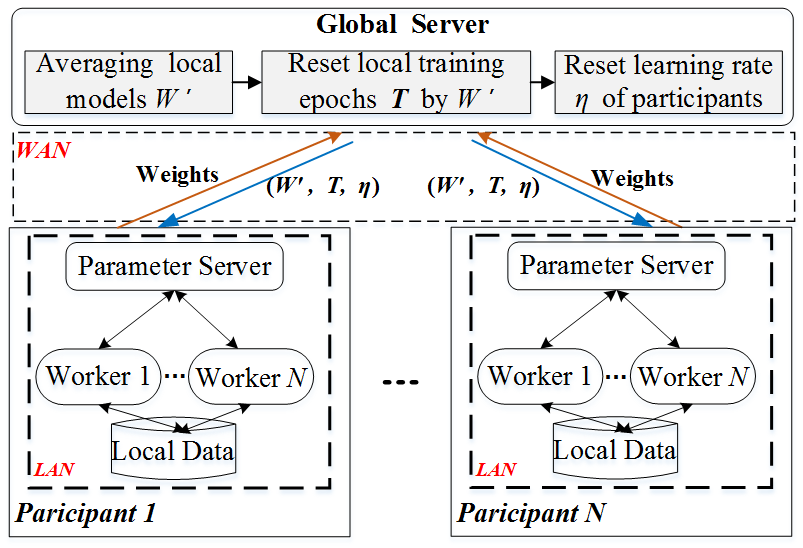}
	\caption{Workflow of co-learning. Assume that the participants are different data centers. Each participant holds an amount of private data and uses the disjoint data to train a local classifier. The local model parameters will be averaged by the global server to formulate the new shared model, which in turn are used for as the starting point for the next round of local training. Besides the new shared model, the global server also updates the number of local training epochs and the learning rate.}
	\label{fig:framework}
\end{figure}

\begin{itemize}

\item{Local training for the participants:}
At the $t$-th iteration of round $i$, participant $k$ updates its local model by using SGD. We refer to one full iteration over all local training data as an epoch. The local model is communicated and aggregated to formulate a shared model after $T_i$ epochs, which is decided dynamically by the global server. Then each participant can initialize its local parameters for the following local training by downloading latest values of the shared model from the global server. During the local training, the participant does not need to exchange the data with other participants. At the iteration $t$ of $i$-th round, the empirical loss of the $k$-th local model is defined as
\begin{equation}
\mathcal{L}(f({\bf w}_k^{i,t},x_k),y_k)=\sum_{m=1}^{m_k}\mathcal{L}(f({\bf w}_k^{i,t},x_{k,m}),y_{k,m}).
\end{equation}

Specifically, participant $k$ updates its local model from ${\bf w}_k^{i,t}$ to ${\bf w}_k^{i,t+1}$ by minimizing the
training loss using SGD.


\item{Model aggregation for the global server:}
Firstly, the global server initializes the shared model parameters and pushes them to all participants. The local training of each participant follows the aforementioned procedures. If one participant $k$ fails to upload its parameters due to network errors or other failures, the global server will restart the local training process of participant $k$. After all $K$ participants finish their updates in the $i$-th round and obtain the parameter ${\bf w}_k^{i}$, the global deep neural network model is updated by taking the average of the $K$ sets of parameters, i.e.,
\begin{equation}
\bar{{\bf w}}^{i}=\frac{1}{K}\sum_{k=1}^{K}{\bf w}_k^{i},
\end{equation}
which is further sent back to the local participants, and set as the initial parameters for the following training. Further, the number of epochs $T_i$ is reset according to the conditions defined in Equations (4). The parameters of the shared model, as well as $T_i$ and $\eta^i$, are sent back to local participants, and used as the starting point for the next round of local training (as can be seen in Figure 1).

\subsection{Cyclical learning rate and increasing local epochs}
To avoid falling into local optima, we employ the cyclical learning rate (CLR) schedule in the training phase of the local participants. Specifically, within the $i$-th communication round, we decay the learning rate with an exponential annealing for each epoch $j$ as follows:

\begin{equation}
\eta^i_j=\eta^{i}\times r^{\frac{j}{T_i}},
\end{equation}

$r$ is the decay rate (in our experiment, $r$ is set as $1/4$), $\eta^{i}$ is the shared learning rate in the $i$th round, used as an initial value to update each participant's local learning rate. It can be updated as $i$ grows. For simplicity, we set $\eta^i$ as a constant value (i.e. 0.01) in this paper.
As mentioned above, the global server has to decide the number of epochs for local participants dynamically, since these values have a significant impact on the accuracy of the shared model. The number of local epochs in the $i$-th round ($T_i$ ) is updated based on the following rules:
\begin{equation}
	\label{eq:alpha}
	T_i =\left\{
	\begin{array}{lcl}
		 T_{0},\;   &if \; i=0,\\
		 2*T_{i-1},\;   &if \; i>0~\&~\frac{|\bar{{\bf w}}^{i}-\bar{{\bf w}}^{i-1}|}{|\bar{{\bf w}}^{i-1}|}\leq\epsilon,\\
		 T_{i-1},\;   &if \; i>0~\&~\frac{|\bar{{\bf w}}^{i}-\bar{{\bf w}}^{i-1}|}{|\bar{{\bf w}}^{i-1}|}>\epsilon,
	\end{array}
	\right.
\end{equation}

where $\epsilon$ is used to control the convergence precision of the shared model parameters. In other words, the number of epochs in each round is increased by a factor of $2$ at every communication round once the change of the shared model parameters is lower than $\epsilon$. The pseudocode of the proposed co-learning is given in Algorithm 1.

\begin{algorithm}
	\caption{co-learning}
	\begin{algorithmic}
		\State initialize ${w^0}$, $\eta^0$ and $T_0$
		\For {each round $i$ = 0, 1, 2, ..., N}
		\State reset $T_i$ according to the Equation (4)
		\State send ${w}^{i}$, $\eta^i$ and $T_{i}$ to participants
		\For{ each participant $k$ $\in$ K \textbf{parallel} }
        \For {local epoch $j$ from 1 to $T_i$}
        \State update $\eta^i_j$ according to the Equation (3)
		\State {${w^{i}_{k}} \gets $ localSGD(${w}^{i}$, $\eta^i_j$)}
		\EndFor
		\State upload ${w}^{i}_{k}$ to server
		\EndFor
		\State {${w}^{i+1} \gets \frac{1}{K}\sum_{k=1}^{K}{w}_k^{i}$}
		\EndFor
	\end{algorithmic}
\end{algorithm}

\end{itemize}

\subsection{Ablation study on CLR and ILE}
In this part, we perform a thorough ablation study to highlight the benefits of cyclical learning rate (CLR) and increasing local epochs (ILE) on model averaging. We also employ the exponential learning rate (ELR, i.e. non-cyclical learning rate) and fixed local epochs (FLE) for the quantitative comparison.

We run experiments on the CIFAR-10 dataset, which consists of 10 classes 32$\times$32 images with three channels. 50,000 training images are partitioned into five disjoint subsets, which are stored in five different data centers, and each containing 10,000 samples. The 10,000 test images are used for the evaluation. The initial values of $T_0$ for the DenseNet-40, ResNet-152, Inception-V4, and Inception-ResNet-V2 models are 5, 5, 20, 5 respectively. The batch size of the experiments was set to 32. 

Using the pairwise combination of (cyclical learning rate (CLR), exponential learning rate (ELR)) and (increasing local epochs (ILE), fixed local epochs (FLE)), Figure 2 shows the accuracy of model averaging method for training DenseNet-40, ResNet-152, Inception-V4 and Inception-ResNet-V2. As can be seen from the figure:

\begin{itemize}
\item The combination of CLR and ILE achieves the highest accuracy on four different network architectures. The results demonstrate that co-learning (CLR+ILE, the red line) tends to generalize better, which indicates the benefits of both cyclical learning rate and increasing local epochs. The reason behind might be that co-learning could converge to flat local optima rather than sharp, isolated optima. Such flat regions are robust to data perturbations as well as perturbations of the parameters, all of which are crucial factors to achieve good generalization.

\item Similar to previous studies using model averaging, the combination of ELR and FLE (the green line) cannot effectively improve the performance of the collaborative learning, and tends to be over-fitting in the training phase. In other words, the performance of the shared model cannot be improved by using model averaging alone without any optimization strategy.

\item Further, ELR+ILE leads to a converged result, however, the CLR+FLE prones to be over-fitting. This indicates the ILE may bring more performance gains than the CLR on the CIFAR-10 dataset, and ILE can increase the diversities between different local models, which consequently derives a better shared model.
\end{itemize}

\begin{figure*}[ht]
	\centering
	\includegraphics[width=0.8\textwidth]{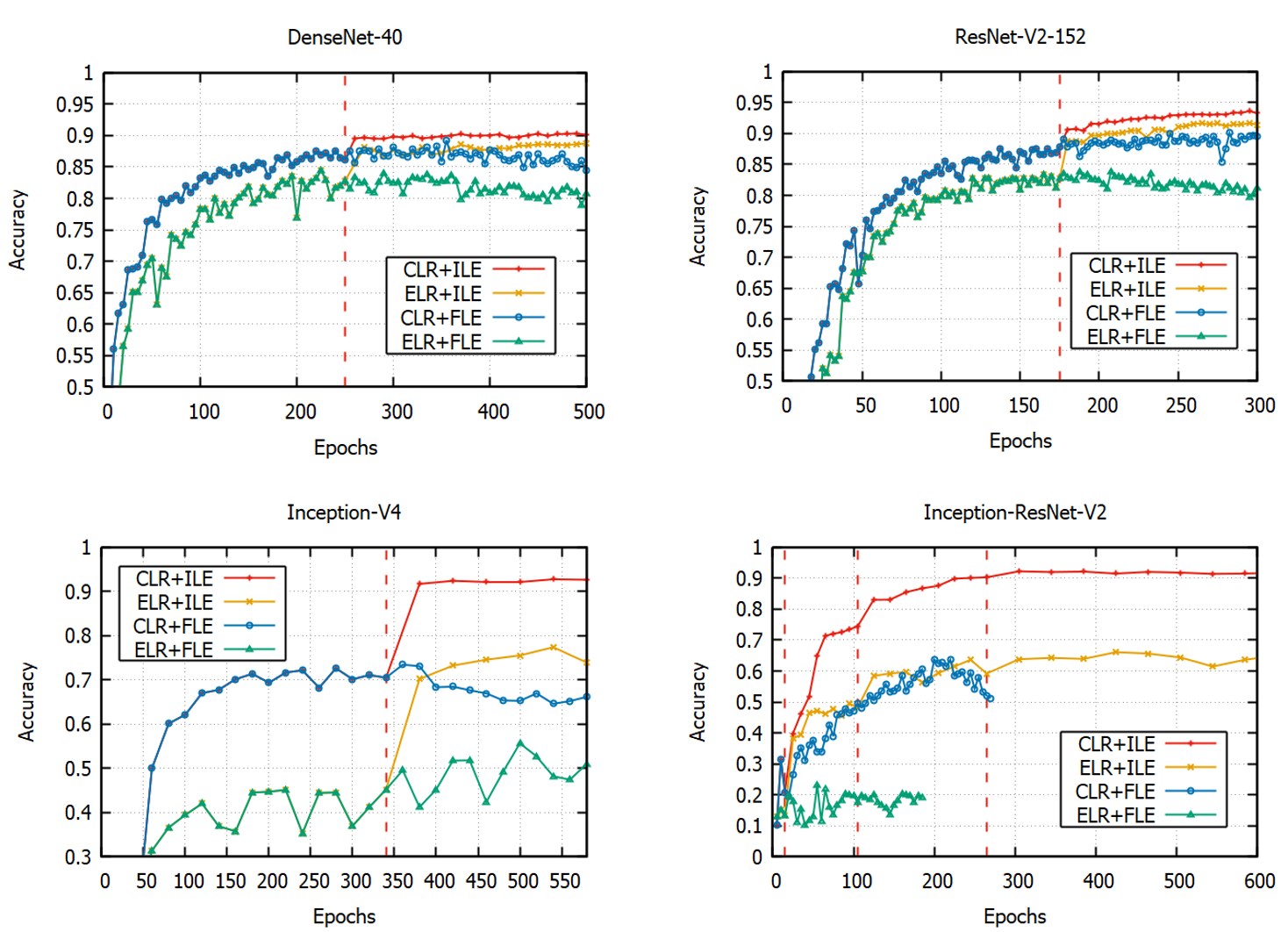}
	\caption{Accuracy on the CIFAR-10 dataset by using different strategies. The employed neural network architectures include: Inception-V4, ResNet, Inception-ResNet, DenseNet. Using the proposed ILE strategy, DenseNet-40, ResNet-152, Inception-V4 enlarges $T_0$ at the 250th, 175th and 340th epoch respectively, while Inception-ResNet-V2 increases $T_0, T_1, T_2$ at the 15th, 105th and 265th epoch, respectively. After the adjustment, the performance of each shared model sees a significant improvement in the following rounds. The FLE strategy in the bottom-right figure (the blue line and green line) experiences an early stop, as it does not boost the performance in the previous rounds.}
	\label{fig:ablation_study}
\end{figure*}

\subsection{Communication cost}
We briefly summarize the communication cost for the proposed co-learning approach. Table 1 exhibits the communication interval and the transferred volume of one model in a round. The $2$nd column reveals the communication interval between a local participant and the global server in a communication round before $T_0$ is increased (i.e. time elapsed between two consecutive model-synchronizations). Specifically, using the CLR+ILE strategy, the communication intervals for different models range from minutes to hours, e.g. 60 minutes for the Inception-V4 and 27.5 minutes for the Inception-ResNet-152. Moreover, if $T$ is enlarged in the following training, the communication interval will be further extended. Take the Inception-V4 as an example, in the $340$th epoch, the number of local epochs $T$ is increased from 20 to 40. Consequently, the communication interval is enlarged from 60 minutes to 120 minutes, which can greatly alleviate the dependence on the WAN bandwidth.

In short, combining the CLR and ILE, the performance of the shared model can be increased, while the communication cost can be reduced. It is also worthwhile to notice that we do not employ the compression technique by which the communication cost can be further decreased.

\begin{table}
	\caption{Stats for using CLR+ILE on different models in a communication round.}
	\label{time-model}
	\centering
	\begin{tabular}{lll}
		\toprule
		Models & \makecell{Comm. interval \\ (min. / $T_0$)}  &  \makecell{Comm. volume \\ (MB)} \\
		\midrule
		DenseNet-40 & 4.5 / 5 & 13 \\
		ResNet-152   & 30 / 5 & 223 \\
		Inception-V4 &  60 / 20 & 168	\\
		Inception-ResNet-V2   & 27.5 / 5 & 218	 \\	
		\bottomrule
	\end{tabular}
\end{table}

\section{Experiments}

\subsection{Experimental Settings}

To demonstrate the effectiveness of co-learning, empirical experiments were conducted on three different tasks: image classification, text classification and audio classification. For image classification, both CIFAR-10 and ImageNet-2014 \cite{russakovsky2015imagenet} were used for the experiments; For text classification, Toxic comment classification dataset was used in the classification tasks; For audio classification, Google speech command data \cite{sainath2015convolutional} and Audio Set \cite{gemmeke2017audio} were employed. Using the proposed co-learning method, different neural network architectures were tested, including state of the art neural networks architectures. We conducted experiments across five geo-distributed data centers in a public cloud, each equipped with a GPU server with four Tesla P40. All kinds of datasets were randomly allocated to 5 participants in an equally distributed manner. All our experiments were implemented in TensoFlow slim. Also, it is worthwhile to notice that all the results were obtained using the average of five repetitive trials of the experiments. The following two groups of experiments were conducted.
\begin{itemize}
    \item It is a common strategy to integrate the training results of each participant by using ensemble learning. In more detail, each participant independently trains its own model, without interacting with other participants during the training process. The average output of each participants model is used as the final prediction. With the CIFAR-10 dataset, accuracy comparison between ensemble-learning and co-learning were carried out on different kinds of network architectures. Besides, training a deep model using the entire dataset in a single data center (denoted as vanilla-learning below) is introduced as a reference for comparison. Except for the two proposed strategies for co-learning, other configuration settings for vanilla-learning are kept the same as the settings of co-learning.

    \item Moreover, to make a quantitative comparison between the data centralized training method and de-centralized one, we conducted comprehensive experiments using vanilla-learning and the proposed co-learning on different kinds of deep network architectures and various types of datasets.
\end{itemize}

\subsection{Ensemble-learning, vanilla-learning and co-learning}
In the following experiment, using the CIFAR-10 dataset, we show the comparison between ensemble-learning, vanilla-learning and co-learning, on five kinds of models (i.e. VGG-19, ResNet-152, Inception-V4, Inception-ResNet-V2, and DenseNet-40). For the vanilla-learning, the exponential learning rate (ELR) is employed. Table \ref{ensemble} illustrates the results. It can be observed that using ensemble-learning, the model accuracy is significantly declined, i.e. nearly 10\% reduction compared with the vanilla-learning. As each participant has only 1/5 disjoint training data, the accuracy of the local model is poor. Consequently, by averaging the outputs of each model after independent local training, it is infeasible to obtain a  competitive performance with the one using vanilla-learning. On the contrary, the accuracy obtained by the co-learning achieves competitive results with comparison to the vanilla-learning. Surprisingly, co-learning on four models (i.e. VGG-19, ResNet-152, Incpeiton-V4 and DenseNet-40) even achieves better performance than the vanilla-learning. These results exhibit again the effectiveness of the cyclical learning rate (CLR) and increasing local epochs (ILE) on model averaging.

\begin{table}
	\caption{CIFAR-10 accuracy comparison between ensemble-learning, vanilla-learning and co-learning.}
	\label{ensemble}
	\centering
	\begin{tabular}{llll}
		\toprule
		\multicolumn{1}{c}{}  & \multicolumn{3}{c}{Accuracy(\%)}\\
		\cmidrule(r){2-4}
		Model    & vanilla  & ensemble & co-learning\\
		\midrule
		VGG-19  & 89.44  & 80.39  & \bf{89.64}\\
		ResNet-152 & 92.64 & 85.4 & \bf{93.51} \\
		Inception-V4 & 91.34  & 83.83  & \bf{92.07}\\
		Inception-ResNet-V2 & \bf{92.86} & 84.7 & 92.83 \\
		DenseNet-40 & 91.35 & 81.24 & \bf{91.43} \\
		\bottomrule
	\end{tabular}
\end{table}

\subsection{Comparison between co-learning and vanilla-learning}

\subsubsection{Image Classification.}
We conduct another image classification experiments on the ImageNet-2014 to further evaluate the generalization accuracy of co-learning, as the classification error on ImageNet is particularly important because many state-of-the-art computer vision problems derive image features or architectures from ImageNet classification models.

In the training phase, we follow standard data augmentation practices: scale and aspect ratio distortions, random crops, and horizontal flips. The batch size is set to 256. Three different state-of-the-art models (VGG, Inception-V4, ResNet-V2-101) are trained, by using both of the co-learning and vanilla-learning approach. Top-1 and Top-5 accuracy rates are reported in Table \ref{imagenet}. We find that the co-learning leads to improved accuracy over vanilla-learning using the same network architecture settings, which illustrates the promising potential of co-learning. This indicates that the co-learning approach can be generically applied to large-scale image classification settings.

\begin{table}
	\caption{Test accuracy of ImageNet-2014 using different models.}
	\label{imagenet}
	\centering
	\begin{tabular}{llll}
		\toprule
		\multicolumn{2}{c}{}  & \multicolumn{2}{c}{Accuracy(\%)}                   \\
		\cmidrule(r){3-4}
		Model                               &             & Top-1 & Top-5   \\
		\midrule
		\multirow{2}{*}{VGG-19}            &  vanilla & 70.41  & 88.12    \\
		&  co-learning  & \bf{70.62}  & \bf{88.7}    \\
		\midrule
		\multirow{2}{*}{Inception-V4}      &  vanilla & 79.16  & 93.82     \\
		&  co-learning  & \bf{79.35}  & \bf{94.28}     \\
		\midrule
		\multirow{2}{*}{ResNet-V2-101}    &  vanilla & 75.66  & 92.28     \\
		&  co-learning  & \bf{75.85}  & \bf{92.39}     \\		
		\bottomrule
	\end{tabular}
\end{table}

\subsubsection{Text Classification.}
We also run experiments on a large-scale toxic comments classification task to demonstrate the effectiveness of co-learning on a natural language processing problem. In more detail, the training dataset consists of 159,571 Wikipedia comments, which have been labeled by human raters for toxic behavior, while 153,164 records are used for the evaluation. The types of toxicity are: toxic, severe toxic, obscene, threat, insult, identity hate. In the training stage, the training dataset is randomly partitioned into 5 participants. Each contains equal-size disjoint examples, which are stored in the different data center.

For the classification, the employed models include LSTM \cite{greff2017lstm} and Capsule \cite{hinton2018matrix}. The input embeddings for each word are of dimension 300 (for the pre-trained word vectors, fastText \cite{bojanowski2017enriching} is employed). For LSTM model, we use a bidirectional GRU and the batch size is set to 128 here. For Capsule model, the input is the reshaped embedding vectors, while the second layer is a primary capsule layer with strides of 1. This layer consists of 32 ``Component Capsules'' with a dimension of 8. Final capsule layer includes 6 capsules, refereed to as ``Class Capsules'', one for each type of toxicity. The dimension of these capsules is 16.

For the evaluation, the mean column-wise ROC AUC is used. As can be been from the Table \ref{Toxic}, the co-learning improves the accuracy with comparison to the vanilla-learning. The experimental results suggest that our method is practically applicable to the large-scale text classification task.

\begin{table}
	\caption{Multi-class AUC on toxic comment classification challenge dataset. }
	\label{Toxic}
	\centering
	\begin{tabular}{lll}
		\toprule
		\multicolumn{1}{c}{}  & \multicolumn{2}{c}{Multi-class AUC(\%)}                   \\
		\cmidrule(r){2-3}
		Model    & vanilla & co-learning \\
		\midrule
		LSTM     & 98.52 & \bf{98.79}     \\
		Capsule  & 98.32  & \bf{98.75}      \\
		\bottomrule
	\end{tabular}
\end{table}

\subsubsection{Audio Classification.}
Next, we conduct experiments on the audio classification task. Two different datasets are used: Google commands dataset and Audio Set.

\begin{itemize}
\item{Google Command Recognition.} Google commands dataset contains 65,000 utterance, in which each audio is about one second long and belongs to one out of 30 classes. The voice commands include classes, such as left, right, yes, no. To process the utterances, we first calculate the log Mel spectrograms from the original raw audio signal at a sample rate of 16 kHz. The model architecture consists of two convolutional layers followed by two fully connected layers and then a softmax layer for classification. While this model is not the state-of-the-art, it is sufficient for our needs, as our goal is to the quantitative study, not achieve the best possible accuracy on this task. Table \ref{TensorFlow} gives the recognition accuracy of the co-learning, and vanilla-learning. As can be seen from the table, nearly the same accuracy can be achieved using the co-learning.
		
\begin{table}
		\caption{TensorFlow speech commands recognition}
		\label{TensorFlow}
		\centering
		\begin{tabular}{lll}
			\multicolumn{2}{c}{}                  \\
			\cmidrule(r){1-3}
			Method     & \makecell{Validation \\ accuracy (\%)}  &  \makecell{Test \\ accuracy (\%) }\\
			\midrule
			vanilla	     & 93.1		       & \bf{93.3}     \\
			co-learning        & \bf{93.3}			   & 93.2    \\
			\bottomrule
		\end{tabular}
\end{table}

\item{Audio event classification using Audio set.} 	To make a quantitative comparison between the co-learning and the vanilla-learning, large-scale audio event classification experiments are conducted. Audio Set consists of a large ontology of 632 sound event classes and a collection of 2 million human-labeled sound clips (mostly 10-second length) drawn from 2 million YouTube videos.
	
Each audio recording feature has 240 frames by 64 mel frequency channels, which are employed as the input for different architectures. The convolutional recurrent neural networks (CRNN) are adopted for the classification task. Specifically, one bi-directional gated recurrent neural network with 128 units is used. Instead of applying a single-level attention model after the fully connected neural network, multiple attention modules \cite{yu2018multi} can be applied after intermediate layers as well. The batch size is set to 128 for different network architectures. Table \ref{sample-table} summarizes the results of different network architectures. Overall, the accuracy is similar by using the co-learning and the vanilla-learning. The result demonstrates the general applicability of our method on audio datasets.
		
\begin{table}
		\caption{Audio Set classification task using a single / multi data center(s). AP represents result of CRNN with average pooling, MP for CRNN with max pooling, SA for CRNN with single attention and MA for CRNN with multi-attention.}
		\label{sample-table}
		\centering
		\begin{tabular}{llll}
			\toprule
			\multicolumn{4}{c}{vanilla / co-learning}                   \\
			\cmidrule(r){1-4}
			Models    			  & MAP    & AUC	& d-prime \\
			\midrule
			AP       & $\bf{0.300}$ / 0.299  & $\bf{0.964}$ / 0.962  & $\bf{2.536}$ / 2.506		\\
			MP           & 0.292 / 0.292  & $\bf{0.960}$ / 0.959  & $\bf{2.471}$ / 2.456		\\
			SA   & 0.337 / 0.337  & $\bf{0.968}$ / 0.966  & $\bf{2.612}$ / 2.574		\\
			MA   & $\bf{0.357}$ / 0.352  & 0.968 / 0.968  & $\bf{2.621}$ / 2.618		\\
			\bottomrule
		\end{tabular}
	\end{table}
\end{itemize}

\section{Discussion and Conclusion}

In this paper, we present co-learning, a novel collaborative deep learning approach, for training deep models on disjoint multi-party datasets. Extensive experiments are conducted on different types of data, including image, text, and audio, with the goal to demonstrate the effectiveness of co-learning both quantitatively and qualitatively. All the experiments demonstrate that co-learning method can provide competitive (sometimes, even better) performance, with comparison to the data centralized learning.

The experiments also indicate the benefit of both cyclical learning rate and enlarging local training epoch strategies. The reason behind might be that co-learning could converge to flat local optima rather than sharp, isolated local optima. Such flat regions are robust to data perturbations as well as perturbations of the parameters, all of which are crucial factors to achieve good generalization.

On one hand, by restarting the optimization with a large learning rate, the intrinsic random motion across gradient direction prevents the model from reaching any of the sharp basins along its optimization path, which allows the model to find a better local optima. In this way, although the performance temporarily suffers when the learning rate cycle is restarted, the performance eventually surpasses the previous cycle after annealing the learning rate. On the other hand, by increasing the number of local epoch in the iterations, each local model could do large steps in the parameter space to get diverse networks. Thus, it is expected to achieve better possible accuracy on its local datasets. Moreover, the increasing local epochs leads to add the diversities between different local models, which can be averaged to get a better shared model.

In brief, our co-learning method offers a solution for collaborative deep learning in the context of multi-parties data. Future work includes the practical privacy mechanism, secured multi-party computation in the co-learning framework.

\section{Acknowledgments}
This work was supported by the National Grand R\&D Plan(Grant No. 2016YFB1000101).
\bibliographystyle{aaai}  
\bibliography{reference}

\end{document}